# Sensing Danger: Innate Immunology for Intrusion Detection

Uwe Aickelin and Julie Greensmith
School of Computer Science
University of Nottingham
NG8 1BB
UK

E-mail uxa, jqg@cs.nott.ac.uk

**Abstract**

The immune system provides an ideal metaphor for anomaly detection in general and computer security in particular. Based on this idea, artificial immune systems have been used for a number of years for intrusion detection, unfortunately so far with little success. However, these previous systems were largely based on immunological theory from the 1970s and 1980s and over the last decade our understanding of immunological processes has vastly improved. In this paper we present two new immune inspired algorithms based on the latest immunological discoveries, such as the behaviour of Dendritic Cells. The resultant algorithms are applied to real world intrusion problems and show encouraging results. Overall, we believe there is a bright future for these next generation artificial immune algorithms.

**Introduction**

Artificial Immune Systems (AIS) provide an ideal inspiration for Computer Security in general and Intrusion Detection Systems (IDS) in particular. AIS have been successfully applied to a number of problem domains including fault tolerance, data mining and computer security (Kim 2007). The algorithms central to many AISs and in particular those applied to computer security (Hofmeyr 2000) were based on relatively simplistic models of T-cells, such as the Negative Selection Algorithm. Unfortunately, these simple algorithms have been shown to scale poorly and produce low detection rates often with excessive false positive rates. This effect has been proven both experimentally (Kim 2001) and theoretically (Stibor 2006). A general overview of this area of research is given in Kim *et al.* (Kim 2007).

Yet, the biological immune system is a very effective anomaly detector- surely we should be able to build AISs which do the same? This is the puzzle that started our work some five years ago through the so called 'Danger Project' (EPSRC GR/S47809/01). Our conclusion then was that AIS algorithms to date have largely been inspired by the adaptive immune system and by biologically-naive models. We hypothesised that new research in AIS needs to focus on building more biologically-realistic algorithms which are inspired by both the innate and adaptive immune systems. This is as both of these components of the human immune system are vital to the high level of protection the immune system provides to the host. This paper gives a summary of the work that has happened so far towards the goal of discovering the 'missing link' between AIS and IDS.

The aim of this paper is to provide an overview of two algorithms developed as part of the Danger Project - the Dendritic Cell Algorithm (DCA) and the Toll-like Receptor algorithm (TLR). Both algorithms were developed in parallel within the scope of the project. Whilst using similar immunological concepts for inspiration, both algorithms focus on different aspects of innate immunology to form the basis of the algorithms. In this paper the motivation for the development of both algorithms is provided in the Background section, in addition to relevant immunological context information. This is followed by a summary of the development of both algorithms. Finally we conclude with a qualitative comparison between the two algorithms and suggest future directions for the developed systems.

**Background**

The motivation of the Danger Project was to understand the actual intrusion detection mechanisms employed by the immune system and to capture the essence of these mechanisms through an abstraction process. These abstract models are implemented to form feasible algorithms, capable of performing a useful computational function. In order to achieve this, an in-depth understanding had to be achieved in an emerging concept in immunology known as the 'Danger Theory'. The Danger Theory (Matzinger 1994) states that the immune system is activated upon receipt of molecular signals which indicate damage or stress to the host rather than by pattern matching of 'non-self' versus 'self'.

The Danger Theory is controversial within the world of immunology, as it challenges theories previously thought to be central to the function of the human immune system. The concept of *self non-self discrimination* performed by the immune system has been the cornerstone of immunology since its postulation by Paul Ehrlich in the late 19th Century. The central tenet of immunology is that the immune system responds to the presence of foreign entities (termed non-self) and does not respond to the host (termed self). This discrimination process is achieved through the careful filtering of the *T-cells* of the adaptive immune system, where cells which react to self proteins are deleted. This implies that the immune system consists of a highly tuned population of detector-like T-cells primed only to match and destroy 'non-self' entities, inclusive of viruses, bacteria and other pathogenic microbes.

However, questions have been raised regarding the validity of the self non-self paradigm, due to its inability to explain a number of documented phenomena. For example it cannot explain why autoimmune diseases can occur, where the body responds inappropriately to its own proteins generating natural 'false positives'. Also it does not explain why we do not react to our changing body during pregnancy or to the food in our intestines, not to mention the plethora of 'friendly bacteria' which inhabit our lower intestines, which one can buy and consume nowadays! The Danger Theory postulates that the immune system is not activated upon the detection of non-self entities alone, and does not react to a potential intruder until damage or *danger* is detected. It has been shown that these 'danger signals' are released as a result of *necrotic* cell death which can be caused by pathogenic infection. The presence of danger signals in combination with proteins secreted from the potential infectious agent - termed 'antigen' is required to illicit a defensive response by the human immune system.

Developments in AIS have in some respects paralleled the developments in immunology. The majority of AIS research towards building successful IDS has focused on the theories of self non-self discrimination. From this, arose the development of the Negative Selection Algorithm, which has been used extensively within AIS. The negative selection algorithm is a supervised learning algorithm, consisting of training and testing phases. The goal of the negative selection algorithm is to classify bit-string representations of real-world data (termed 'antigen') as *normal* or *anomalous* (Forrest 1999). To achieve this, detectors are created consisting of a similar bit-string, with which to match the incoming antigen data. During the training phase, the detector population is 'presented' a subset of antigen, taken only from the 'normal' class. Any detector which matches a normal antigen during the training phase is deleted. Once training is complete, the remaining detectors are tuned to respond only to the data items of the nonself or anomalous class as, in theory, all of the self-matching detectors have been eliminated. The system is then presented test data to classify as normal or anomalous.

As with the developments in immunology, similar problems have surfaced within AIS. Early work with the negative selection algorithm looked promising, but as the algorithm was further characterized and applied to larger more complex data, it suffered from the problem of the generation of excessive numbers of false positives (Kim 2001). In addition, the random generation of the detector's bit-strings gave rise to both excessive detector generation and 'holes' in the coverage across the search space. This resulted in significant scaling problems for systems using the negative selection algorithm. Despite many modifications to representation, matching rules and combination with other AIS and evolutionary algorithms, this problem has not been solved. To further criticism, a theoretical proof of the scaling problems and false positive generation was recently performed by Stibor *et al.* (2006).

As with immunology, AIS practitioners started to question the validity of the self-nonself based algorithms. In 2002, Aickelin and Cayzer (2002) proposed that the use of cutting edge immunological principles may hold the key for building successful AIS algorithms. As a result of this notion, the Danger Project was proposed and subsequently instigated as an interdisciplinary research project, involving both a team of practical immunologists and biologically inspired computer scientists.

The Danger Project looked at two aspects of the danger model. Immunologists examined how potential danger signals affected the cells of the immune system. In collaboration with the immunology team, computer scientists researched how the incorporation of the danger model could be used in the improvement of artificial immune systems. This is performed in order to construct improved anomaly detection systems for computer networks. To achieve this, the cells involved in the recognition of danger were examined in considerable detail. These cells are the Dendritic Cells (DCs) of the innate immune system.

As part of our work, we have developed design principles (Twycross 2005) and built a general system and API named *libtissue* within which a number of different artificial immune system algorithms can be implemented (Twycross 2006). Subsequently, we have implemented a range of Danger Theory inspired algorithms, the most advanced ones being the toll-like receptor or TLR algorithm (Twycross 2007) and the DCA (Greensmith 2006). Descriptions of these algorithms follow a brief description of the underlying immunology.

A key requirement to building such immune inspired intrusion detection systems and algorithms had to be an improved understanding of the correlation of signals processed by the DCs of the innate immune system (as depicted in Figure 1), which includes characterisation of danger signals. This is currently a highly active area of research within immunology and is far from fully understood. As part of our project we had to conduct extensive wet lab experiments (Williams 2007) to add to the information derived from the existing literature, as knowledge regarding the behaviour and function of DCs *in vivo* is currently incomplete.

<Insert Figure 1 about here>

It is generally thought that DCs are sensitive to their local environment and are affected by proximal cell death. The immunological part of our project investigated the effects of two types of cell death on DC function. DCs exposed to the signals produced by cells dying in a controlled manner (apoptosis) failed to induce DC maturation and were unable to support the T-cell cloning required for an immune response. However, apoptotic cells co-incubated with necrotic cells significantly suppressed the effects of immune activating signals and attenuated T-cell cloning. Hence our study strongly suggests that apoptosis-induced DC suppression is not an immunological null event, but an active part of the peripheral tolerance mechanism in the body.

Following the above immunological experiments, we concentrated on DCs and T-cells in our computational models, having identified them as the key agents within Danger Theory. DCs collect proteins termed antigen whilst being exposed to environmental signal molecules caused by cell death and other events. The combination of signals determines the DCs pathway to instigate either tolerance or response when interacting with T-cells. T-cells process the information handed over by DCs and then either de-activate or clone and subsequently respond. Therefore our task was to design a computer system that emulates DCs and T-cells in their role as antigen processors with a signal based classification scheme, i.e. achieved trough multi-sensor data fusion.

In an abstract model of DC behaviour, DCs exist in one of three states at any given time: immature; semi-mature and fully mature. The transition between these various DC states is shown in Figure 2. Input signals form four categories, inclusive of PAMPs (pathogen associated molecular patterns), danger signals (Matzinger 1994), safe signals and inflammation. Within the biological system, PAMPs are molecules released exclusively by pathogens; danger signals are released from tissue cells following unplanned necrotic cell death; safe signals are released from normally dying cells as an indicator of healthy tissue; and inflammation is classed as the molecules of an inflammatory response to tissue injury.

<Insert Figure 2 about here>

The receipt of PAMP and danger signals results in the DC producing a mature signal, whereas the receipt of safe signals results in the production of a semi-mature signal. Inflammation acts as a natural amplifier for all other signal categories. Three output signals are produced by DCs in response to signal input: a costimulation signal, and two maturation state signals. The costimulation signal indicates that the DC has collected sufficient signals to make a decision as to its future maturation status. The maturation state signals are subdivided into two opposing signals, mature and semi-mature.

Antigen collected and presented by a DC expressing the mature signal are classed as anomalous by the adaptive immune system and any cell displaying the antigen is destroyed as it is seen as a threat to the body. Conversely, antigen presented by a `semi-mature' DC is seen as part of normal cell function, and the adaptive immune system is tolerised to the presented antigen. This two-directional decision point is the initial intrusion detection component of the human immune system. Due to their importance we will revisit DCs in more detail later in the Dendritic Cell Algorithm section.

**The libtissue system**

The aim of this section is to summarise the implementation of libtissue, a prototype software system for building second generation AISs and applying them to real-world problems. The libtissue software allows researchers to implement AISs as multi-agent systems and analyse the behaviour of these systems when they are applied to real-world problems. This API framework uses the notion of compartmentalisation (Twycross 2006) and tissue to give the system a sense of *embodiment*.

This system has a client/server architecture where an AIS is implemented as part of a libtissue server, and libtissue clients provide input data to the algorithm and response mechanisms which change the state of the monitored system. This client/server architecture separates data collection by the libtissue clients from data processing by the libtissue servers and allows for relatively easy extensibility and testing of algorithms on new data sources. The libtissue framework is coded in C as a Linux shared library with client and server APIs, allowing new antigen and signal sources to be easily added to libtissue servers from a programmatic perspective. AIS algorithms can be compiled and run on other researchers' machines with no modification as libtissue is implemented as a library. Clients and servers can potentially run on separate machines, for example a signal or antigen client may in fact be a remote network monitor.

AISs are implemented within a libtissue server as multi-agent populations of cells. Cells of different types can be created within an environment, called a tissue compartment, along with antigen and signals. The problem to which the algorithm is being applied is represented by libtissue as antigen and external signals. The libtissue clients collect antigen and external signals and pass them to the libtissue server, which makes them available to the AISs. Cells express various types of receptors and producers which allow them to interact with antigen and control other cells through signalling networks. Additionally, libtissue allows data on implemented algorithms to be collected and logged, allowing for experimental analysis of the system. Both algorithms presented in this paper are implemented using this framework.

**The TLR algorithm**

The 'TLR' algorithm is based on innate immune principles and includes abstracted versions of T-cells, naively implemented DCs, negative selection, tissue compartments and lymph nodes. This work encompasses concepts drawn from central tolerance and from the signal model from the infectious nonself theory (Medzhitov 2002). The TLR algorithm is based on two populations of interacting cells, namely DCs and T-cells. The DCs implemented in TLR collect antigen from an antigen store, and process signals. Unlike the DCA, different categories of input signals are not used, with the focus being on the nature of the interactions between DCs and T-cells. In TLR, DCs are created as immature detectors and sample signals and antigen for a finite specified period of time. If the DC receives a signal during antigen collection, it is termed mature, and conversely DCs which did not detect the presence of a signal are termed semi-mature.

Once the DCs lifespan is complete, the cell is transferred to a 'lymph node' in which it is compared against a population of T-cells. The T-cells are assigned sensors termed 'receptors' of the same representation as the antigen presented by the DC population. The T-cell receptor responsible for the matching and interaction with DCs is generated during a training phase. The T-cells exist in two states, namely *naive* and *activated*. A T-cell matching antigen presented by a mature DC is activated whereas T-cells matching semi-mature DCs are removed from the population and deleted. Anomalies are detected and a session is classified as `anomalous' if a population of activated T-cells (one or more) is generated.

The signals used in TLR are referred to as *danger signals,* implying signals which may represent 'damage' or 'danger'. However, the signals used in TLR are more analogous to PAMP signals, indicating an evolutionarily selected signature of intrusion, with a training phase used to simulate this evolutionary process. Additionally, the name 'TLR' is in reference to *toll like receptors*, which are biologically the membrane bound proteins responsible for processing changes in PAMP concentration by DCs. The signals used in the TLR algorithm are binary signals, representing 'signal present' or 'signal not present', compiled during a short training period.

A list of signal values is compiled during a training period, termed as the 'infectious signal list'. This list consists of discrete signal values which, when sensed by a DC, 'activate' the TLRs (i.e. sensors) on the DCs. The infectious signal list is initially generated to cover all values possible for the three signals. The values of signals seen during training are deleted from the infectious signal list, in a similar manner to negative selection, resulting in a list of values which are defined as 'nonself'. During the testing phase when the TLR algorithm is presented with data to classify, DCs mature upon the activation by *any* signal, irrespective of its type or value.

Additionally, the training period is used to negatively select the T-cell receptors present on each cell within the T-cell population. Initially, a population of T-cells are created with receptors generated to match all possible values of input antigen. For example if monitoring processes on a host machine, antigen are represented as system call ID numbers (Twycross 2006) with 256 unique system call IDs whose values are distributed throughout the T-cell population. Whilst using similar concepts as previous negative selection work, this is a smaller number in comparison to the total possible amount of values used for detector generation in most negative selection based approaches. Normal antigen data is used to create a normal antigen profile. This profile is then matched against the entire T-cell population and any T-cell matching a normal antigen is deleted from the population. This results in two populations of tuned cells which perform the anomaly detection.

A summary of the TLR algorithm is given in the forthcoming list, and is represented in Figure 3:

1. Record set of system calls made in training data
2. Record signal values experienced in training data
3. Compute complement set to 1. and 2.
4. Create immature DCs (iDCs) with signal receptors randomly drawn from complement signal set and with antigen receptors randomly drawn from complement system call set
5. Create naïve T-Cells (nTCs) with antigen receptors randomly drawn from complement system call set
6. iDCs are continually exposed to and sample signals and antigen respectively
7. If during its lifetime an iDC's signal receptor matches a signal it becomes a mature DC (mDC) and migrates
8. If not migrated at the end of its lifetime an iDC becomes a semi-mature DC (smDC) at the end of its lifetime and migrates
9. Migrated smDCs and mDCs present their antigen and try and match nTCs
10. If an mDC expresses an antigen that matches an nTC receptor than this turns the nTC into an activated T-Cell (aTC) and we have an anomaly
11. If an smDC expresses an antigen that matches an nTC receptor than this kills the nTC to reduce false positives

12. Migrated smDCs and mDCs and killed nTCs are replaced with new cells as per 4. / 5.

<Insert Figure 3 about here>

The TLR algorithm has been evaluated on a system call anomaly detection problem. A process, in this case an FTP server (including child processes) is monitored and the system calls made by these processes are gathered to be used as antigen. Memory, file and socket resource usage statistics are also gathered and used as external signals. Signal receptors on DCs are activated by certain external signal values and antigen receptors on naive T-cells are activated by certain antigen values. In order to determine which signal and antigen values activate these receptors TLR needs to be provided with a set of training data consisting of a sample of normal instances only.

In the case of antigen receptors a new set of permissible T-cell receptor values is created by removing all antigens observed in the training set from the set of all possible antigen values (around 350 in the case of system call numbers). In a similar way, signal receptors are only activated by signal levels not seen in the training set. However, unlike antigen receptors, signal receptors are not specific for one particular value, but rather any value not seen in the training set.

In order to test TLR we used the publicly available autowux exploit autowux-cert. This exploit levers a format string vulnerability, in this case related to the SITE EXEC FTP command, in order to obtain by default a remote root shell on the server. It has been seen in the wild in manual attacks and automated attacks such as the Ramen worm (SANS 2001). We also performed an FTP bounce scan attack (Hobbit 1995) in which an attacker used an FTP server as an intermediary to perform a network scan and hide the IP address of their machine. Both attack types are mixed in with a large number of normal ftp sessions captured from external data.

On the above problem, the TLR algorithm achieves false positives rates of 0.15 and true positive rates of 0.75. Here we are assuming an equal cost for false and true positives. The TLR algorithm, which was unoptimised, used around 10% of the CPU resources (Athlon XP 2600+, 1 GB RAM) and never more than 8% of the memory resources on the test machine. Generally, CPU usage was only a few percent as cell levels were maintained at a low level during normal usage. Overall, TLR was able to detect anomalous behaviour resulting from attacks with a high true positive and low false negative rate (Twycross 2007) comparable to other state-of-the-art approaches, e.g. Mutz (2006).

**The Dendritic Cell Algorithm (DCA)**

The DCA is based on an abstract model of DC behaviour, initially presented in Greensmith (2005). In nature, DCs perform the function of antigen presentation, where debris found in tissue are collected by DCs, processed to form antigen and presented to the adaptive immune system in combination with *context* information. The context information is derived through the DCs processing of various signals, found in the tissue at the time of antigen collection (Lutz 2002). As a computational technique, the DCA performs correlation of context, derived from the processing of a set of input signals, with antigen - the data to be correlated. This is based on the premise that 'suspects' in the form of antigen can be paired with 'evidence' in the form of signals to identify potential sources of anomaly or intrusion. A general overview of the DCA is provided in this section with a formal description of this algorithm given in Greensmith *et al.* (2006).

As with TLR, the DCA is a population based algorithm, but unlike TLR consists of DC agents alone. The DCA is implemented using the libtissue framework to facilitate the creation and update of cells and tissue attributes. A graphical representation of the DCA is presented in Figure 4. The algorithm processes two input streams consisting of signals and antigen. The signal stream contains a specified number of input signals, which are pre-normalised and categorised as PAMP, danger signal, safe signal or inflammation. A storage facility for incoming signals and antigen is provided and forms the 'tissue' for the DCs. The DCA can be described on two levels: firstly at the level of an individual DC and secondly at the level of the DC population.

<Insert Figure 4 about here>

Each individual DC in the population is a data-fusion agent. As with the natural system, DCs exist in one of three states - immature, semi-mature or mature. Upon creation and initialisation, the cell assigned to the immature state. In this state the cell performs information processing. Cells are updated at regular intervals, where the DC in question copies the values found in the signal matrix to its own internal signal storage. Here the DC has the chance to collect and internalise a specified number of antigen data.

An abstraction of the semantics of the natural signals is used to form a schema for the signal pre-categorisation. This categorisation is based on the following general principles:

- **PAMPs**: Pathogenic associated molecular patterns are proteins expressed exclusively by bacteria, which can be detected by DCs and result in immune activation. The presence of PAMPS usually indicates an anomalous situation.
- **Danger signals**: Signals produced as a result of unplanned necrotic cell death. On damage to a cell, the chaotic breakdown of internal components forms danger signals which accumulate in tissue. DCs are sensitive to changes in danger signal concentration. The presence of danger signals may or may not indicate an anomalous situation, however the probability of an anomaly is higher than under normal circumstances.
- **Safe signals**: Signals produced via the process of normal cell death. Cells must die for regulatory reasons, and the tightly controlled process results in the release of various signals into the tissue. These `safe signals' result in immune suppression. The presence of safe signals almost certainly indicates that no anomalies are present.

- **Inflammation**: Various immune-stimulating molecules can be released as a result of injury. Inflammatory signals and the process of inflammation are not enough to stimulate DCs alone, but can amplify the effects of the other three categories of signal. It is not possible to say whether an anomaly is more or less likely if inflammatory signals are present. However, their presence amplifies the above three signals.

The collected input signals are processed to form cumulative output signals through the use of three weighted sum equations. The mechanism of signal processing is described in detail in Greensmith *et al.* (2007a) and is represented graphically in Figure 5. Incrementing the *costimulation* (CSM) output signal is an important feature of the algorithm, as it provides a limit to the time spent by the DC sampling the data in the 'tissue'. Upon initialisation, each DC is assigned what is termed a *migration threshold.* Upon recalculation of the output signals, the achieved value of the CSM output signal as compared to the cell's migration threshold value. If the value of the CSM signal is greater than the migration threshold, then the DC ceases sampling signals and antigen and is then transferred to a separate compartment and is assigned a new state (semi-mature or mature). Once a cell is removed from the population it is replaced immediately by another cell, fixing the population size at a static level.

<Insert Figure 5 about here>

If the cell does not exceed its migration threshold, it continues sampling and the output signals accumulate. In implementations of the DCA so far, each DC is assigned a random number (within a specified range) for the migration threshold. This ensures that across the population, the DCs sample signals and antigen over different time windows. We believe this to add an element of robustness to the system.

The remaining two output signals of the DC are assessed once this threshold is exceeded. At this point the value of the semi-mature output signal is compared to the value of the mature output signal. The context of the cell is assigned as that of the greatest output signal value. For example, if during its lifespan, the DC had experienced predominantly safe signals, then the value of the semi-mature output signal would be higher than that of the mature output signal. Therefore the state of the cell is assigned as *semi-mature.* Conversely, if the cell experienced predominantly PAMP and danger signals, then the cell is assigned to the *mature* state. In the event that both output signals are equal, the semi-mature context is assigned.

Following the assignment of the DC's context, it then presents all antigens that it has collected over its lifespan. The presented antigen are accompanied by the context of the cell (0 for semi-mature and 1 for mature) and are recorded. This information is used following the processing of all input data to calculate the anomaly coefficient for each type of antigen. A state diagram is presented in Figure 6 which represents this process.

At the population/system level, three events occur namely update of tissue antigen, update of tissue signals and update of DC population. These three processes occur asynchronously. Antigen are updated from the system on an event driven basis i.e. when it is available from the underlying data, it is fed into the tissue's antigen storage. The incoming signals are pre-normalised by a signal collection demon program and are sent to the tissue's signal matrix at a consistent rate. The population of cells is also updated at a specified consistent rate. In our current research in anomaly detection, both cells and signals are updated once per second. This update procedure continues until all data is processed.

Upon completion of data processing the presented antigen-plus context values are analysed to form *MCAV anomaly coefficient values*. The term 'MCAV' refers to the 'mature context antigen value' and is a metric of the proportion of times a particular type of antigen is presented in the mature context. By antigen 'type' we refer to the fact that the antigen data supplied to the system consists of groups of antigen of identical values. For example, if computer programs were monitored as antigen, each time the program performed an operation, an antigen would be generated with the value of the program's ID number. The MCAV calculation returns a coefficient value between 0 and 1. Values closer to 1 indicate that the antigen type has a higher probability of being anomalous. This completes the pairing of 'suspect' antigen with 'evidence' from signals, based on the consensus opinion of the DC population over time.

<Insert Figure 6 about here>

Experimentation with the DCA (Greensmith 2005) has shown that it is suitable for the detection of anomalies in time-dependent data, where the input data consists of a set of input signals to be fused together and ultimately correlated with antigen, for which there are multiple types. To demonstrate this ability, the DCA is applied to the detection of outbound ICMP 'ping' based port scans (Greensmith 2007), with experiments performed in both real-time and offline scenarios.

In these experiments the DCA is used to detect which process is responsible for the invocation of an *nmap* port scan. To perform this detection, process IDs are logged each time an individual process makes a system call. This process ID forms an antigen for the DCA to analyse. Unlike previous AIS approaches to intrusion detection, the DCA does not perform pattern matching on the value of the antigen. Instead the classification of antigen is based on the processing of signals which are found in the system at the time of antigen collection.

In the case of the ping scan investigation, antigen is derived from the process IDs of system calls invoked by running programs, with the aim of identifying an anomalous port scanning program. In this experiment three out of the four input signal categories are used and include the following:

- **PAMPs**: The number of ICMP destination unreachable errors per second. High values indicate likely anomalous behaviour.
- **Danger Signal**: The number of network packets sent per second. Low values of this signal may not be anomalous, giving a high value a moderate confidence of indicating abnormality.

- **Safe Signal**: The inverse rate of change of packet sending. A measure which increases value in conjunction with observed normal behaviour. This is a confident indicator of normal, predictable or steady-state system behaviour. This signal is used to counteract the effects of PAMPs and danger signals, achieved through the negative weight in the signal processing of the DCs.
- **Inflammation:** does not feature in this experiment as no suitable signal was available.

It is shown that the DCA can discriminate between the nmap scan and other co-occurring normal processes, such as the transfer of a large file, with high rates of true positives and low rates of false positives. In Figure 7, the results for the detection of a scan process are shown. Higher MCAV coefficients are derived for the two anomalous processes (nmap and pts) than for the two normal processes (sshd and bash). Upon application of a threshold to the MCAVs at 0.5 all processes are correctly classified.

<Insert Figure 7 about here>

A full sensitivity analysis of the system has been performed (Greensmith 2007a), and the DCA is shown to be robust to changes in the mappings of the signal categories and system parameters, including variation of the number of DCs created and variation of the antigen vector size. Variation is also performed examining the values of the weights used in the signal processing. This investigation shows that changing the values of the weights has no significant impact on the rate of true positives, but incorrectly chosen weight values can increase the observable rate of false positives.

In addition to the application of the DCA to ping scan detection, the algorithm has been applied to a static machine learning dataset, and to the detection of prolonged and more complex port scans, when applied to the detection of SYN scans (Greensmith 2007b). A direct comparison between DCA and TLR is performed in Greensmith (2007c), which showed good detection rates for both algorithms. The DCA has found uses within the detection of misbehaviour in sensor networks (Kim 2006) and for anomalous object detection within mobile robotics (Oates 2007).

**Discussion and Conclusions**

In this paper we have presented the development and application of two algorithms based on the Danger Theory, namely the TLR algorithm and the DCA. Both algorithms employ abstract concepts inspired by innate immunology. In particular, models of DCs are used and abstract computational implementations of these models form the cornerstone of both algorithms. DCs are crucial to the protection provided by the natural human immune system and therefore we believe these cells to have an important role in AIS. The models of DC behaviour present in both algorithms are derived not only from published literature, but also from information produced by collaboration with practical immunologists, as part of the Danger Project. Aside from the focus on DCs, this work also shows that the incorporation of aspects of the innate immune system can be beneficial for the development of AISs. As a secondary outcome, the *libtissue* framework is validated as a feasible API system for the purpose of implementing agent-based AIS algorithms. This is shown through the implementation of both the DCA and TLR within this framework.

TLR has shown to give an improved performance over negative selection on various computer security datasets. This includes a marked reduction in false positives in comparison to a pure negative selection based approach. While these detection rates are promising, it is not possible to make general observations about the performance of TLR as it is difficult to compare to other anomaly detection algorithms due to its specific data requirements. This includes the need in TLR for two databases of input data, one for signals and one for antigen. Standard anomaly detection approaches do not function in this manner and thus many standard IDS datasets are difficult use with TLR

One of the central features of TLR is its requirement for training data. Unlike other AIS approaches, TLR performs training on multiple types of cell agent which appears to add an extra element of tolerance to the generation of false positive errors. However, as with many supervised learning algorithms, training data can be difficult to collect. It is not trivial to produce a *good* training set which does not contain any sources of anomaly. Additionally it is difficult to assess as to what length the training data should be and what diversity of situations can be used to imply normal.

The DCA has a greater reliance on the signal processing aspect by using multiple signal models. The DCA does not require training period, but instead uses expert knowledge to assign input signals to the appropriate category. The DCA also performs anomaly detection with a relatively low rate of false positive errors. However as with TLR, it is difficult to perform direct comparisons with more standardized approaches, as the DCA has similar antigen and signal input data requirements. Nevertheless, the DCA has shown promising results in our own port scanning experiments, and is starting to show good results across a range of problem domains, including sensor networks and mobile robotics.

There are obvious similarities between the two algorithms. The both perform a type of temporal correlation between signals and antigen. Both algorithms contain the concept of 'tissue' and both consist of DCs performing a computational task. Upon closer inspection of the two approaches, marked differences can be seen. TLR defines interactions between both T-cells and DCs. This is coupled with the training period for the selection of both signals and antigen. This is inherently more complex than the single cell and multiple signal model employed by the DCA. As a result, the DCA relies on fewer tuneable parameters, and has been shown in Greensmith *et al* (2007a) to be robust to changes in the majority of these parameters, inclusive of the weight values used in the signal processing mechanism. Additionally, the DCA does not require a training period, which avoids the potential disadvantages discussed above.

In conclusion, the DCA and TLR are proof that it is indeed possible to build feasible AIS algorithms based on the principles of Danger Theory. While TLR is a good anomaly detection algorithm, it is inherently complex. This makes it difficult to understand theoretically and troublesome to characterise experimentally. The DCA is not a simple algorithm, but it is less complex than TLR, primarily due to its single cell type, parameter robustness and its dispensation with the need for training data.

In our future work we intend to further develop the DCA in a number of ways. Future investigation with the DCA may involve thorough benchmarking with standard techniques, a theoretical investigation of the algorithm, and investigation of methods for automated signal selection. Some intended future applications of the DCA may include other time-dependent data, such as further work with port scans, earthquake and medical imaging data.

See http://www.dangertheory.com for a full list of our papers, test data, the libtissue system and both DCA and TLR algorithms.

**Acknowledgments**
This project is supported by the EPSRC, grant number GR/S47809/01, UCL, UWE and HPLabs.

**Biographies**

Uwe Aickelin received a Management Science degree from the University of Mannheim, Germany, in 1996 and a European Master and PhD in Management Science from the University of Wales, Swansea, UK, in 1996 and 1999, respectively.

Immediately following his PhD, he joined the University of the West of England in Bristol, where he worked for three years in the Mathematics Department as a lecturer in Operational Research. In 2002, he accepted a lectureship in Computer Science at the University of Bradford, mainly focusing on computer security. Since 2003 he works for the University of Nottingham in the School of Computer Science where he is now a Reader in Computer Science and Director of the Inter-disciplinary Optimisation Laboratory.

Dr Aickelin currently holds an EPSRC Advanced Fellowship focusing on Artificial Immune Systems, anomaly detection and mathematical modelling. In total, he has been awarded over £2 million EPSRC research funding as Principal Investigator (including an Adventure Grant and two IDEAS Factory projects) on topics including Artificial Immune Systems, Danger Theory, Computer Security, Robotics and Agent Based Simulation. Dr Aickelin is an Associate Editor of the IEEE Transactions on Evolutionary Computation, the Assistant Editor of the Journal of the Operational Research Society and an Editorial Board member of Evolutionary Intelligence.

Julie Greensmith is a Post-doctoral researcher at the University of Nottingham. She gained a BSc in Pharmacology from the University of Leeds, UK in 2002 and a MSc in Multidisciplinary Informatics in 2003, also from the University of Leeds. Following a brief spell in industry working for HP Labs, Bristol, she completed a PhD in Computer Science at the University of Nottingham in 2007.

It was during this time as an MSc student that she became interested in Artificial Immune Systems, which led onto her PhD research, under the supervision of Dr Uwe Aickelin. Her research was performed through an interdisciplinary project which aimed to improve Artificial Immune Systems through the incorporation of a principle known as the Danger Theory. As part of this work, she has developed a means of intrusion detection which is based on an algorithm abstracted from the behaviour of the dendritic cells of the human immune system. As a Post-doctoral researcher, her work now involves extending this algorithm, producing a formal analysis of its behaviour and applying it to real-time problems including earthquake prediction and numerous computer security applications. This research is conducted through Dr Uwe Aickelin's EPSRC Advanced Fellowship which focuses on Artificial Immune Systems, anomaly detection and mathematical modelling.

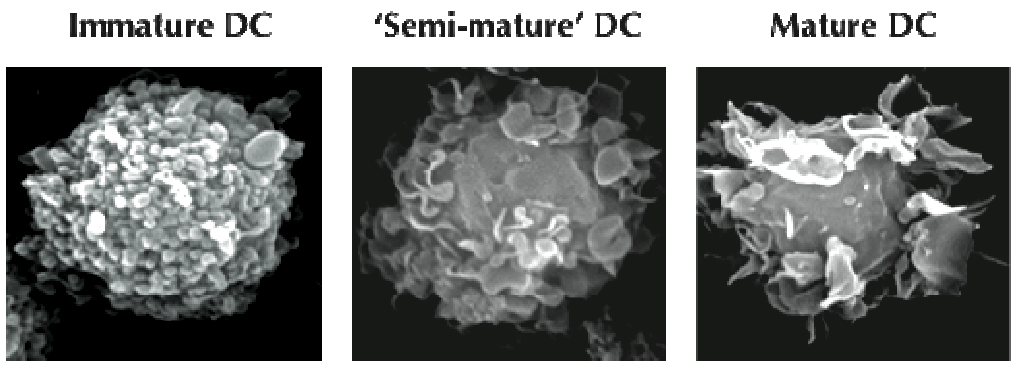

Figure 1: Dendritic Cells under the microscope.

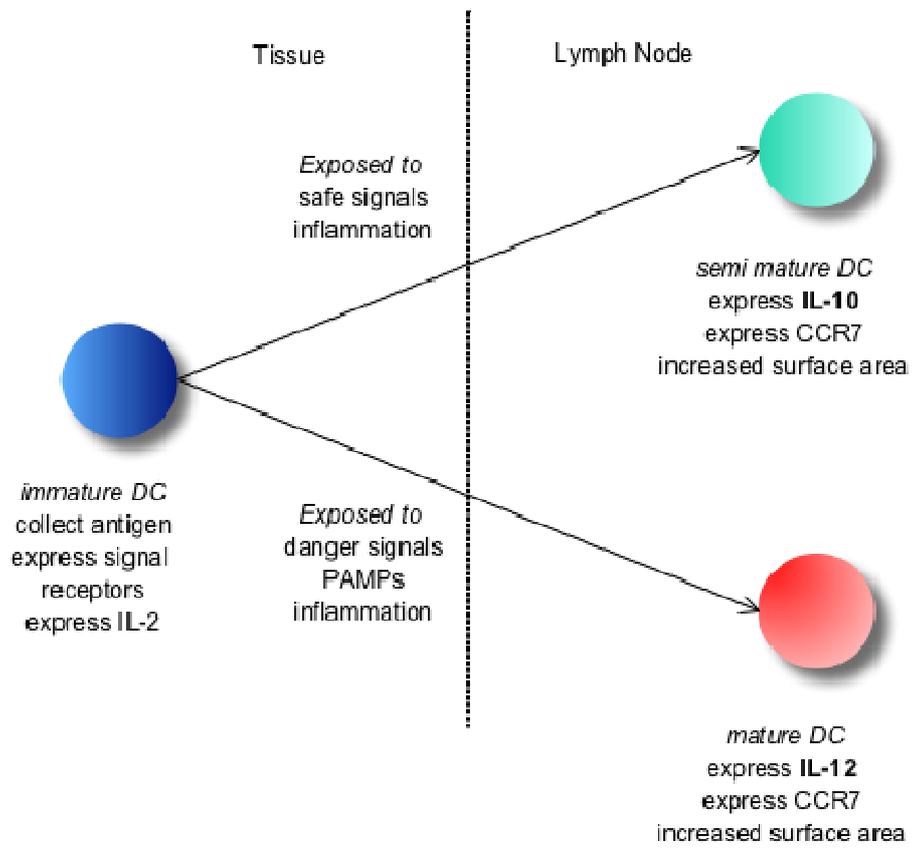

Figure 2: The development of Dendritic Cells

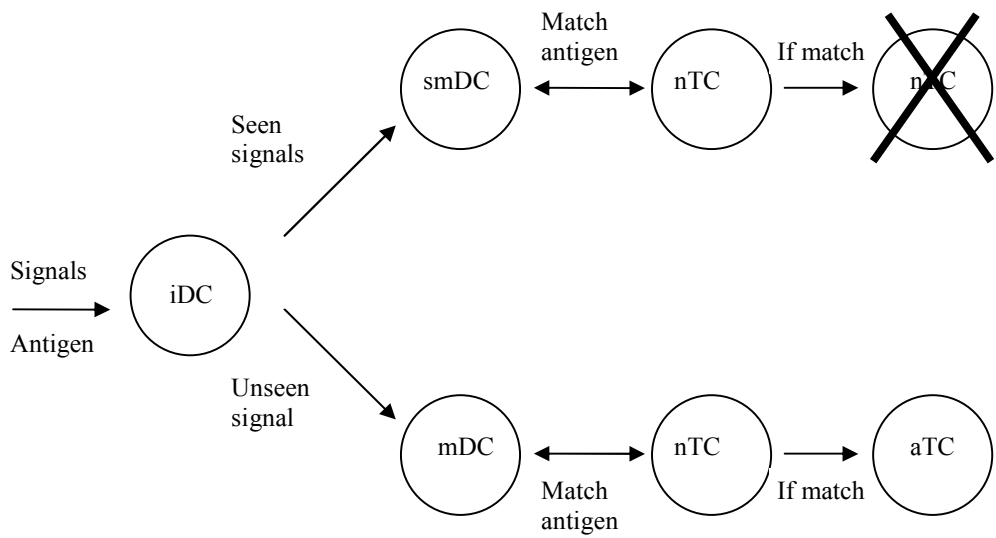

Figure 3: Schematic overview of the TLR algorithm.

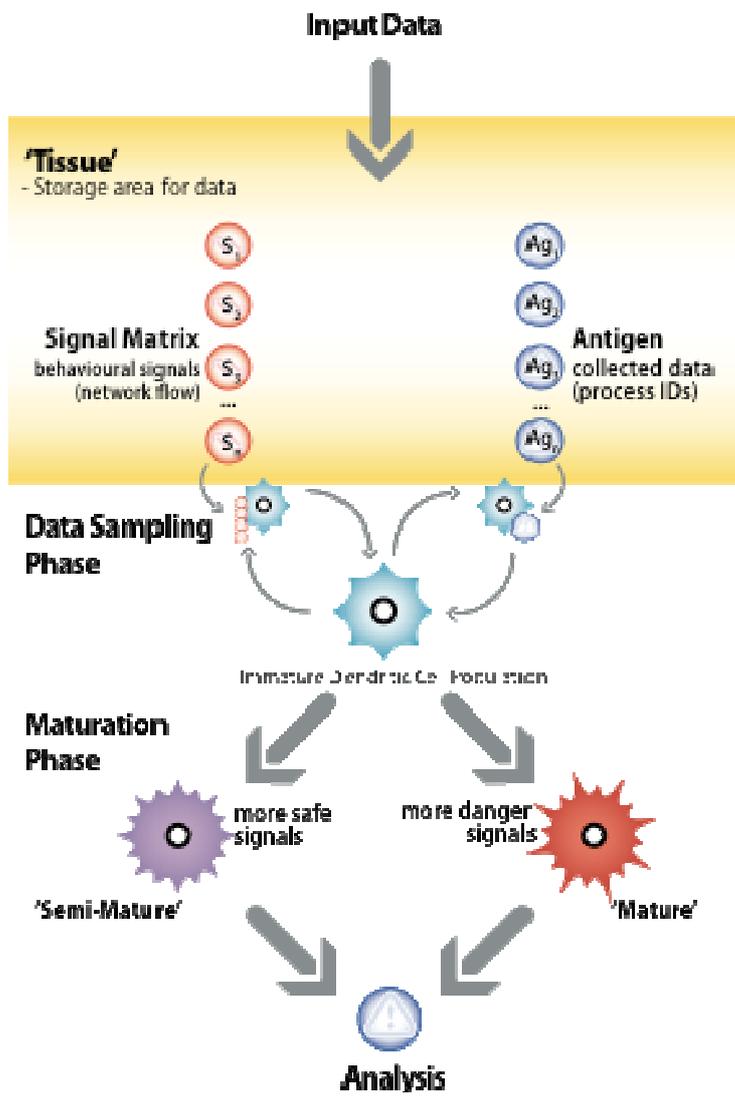

Figure 4: Schematic overview of the DCA.

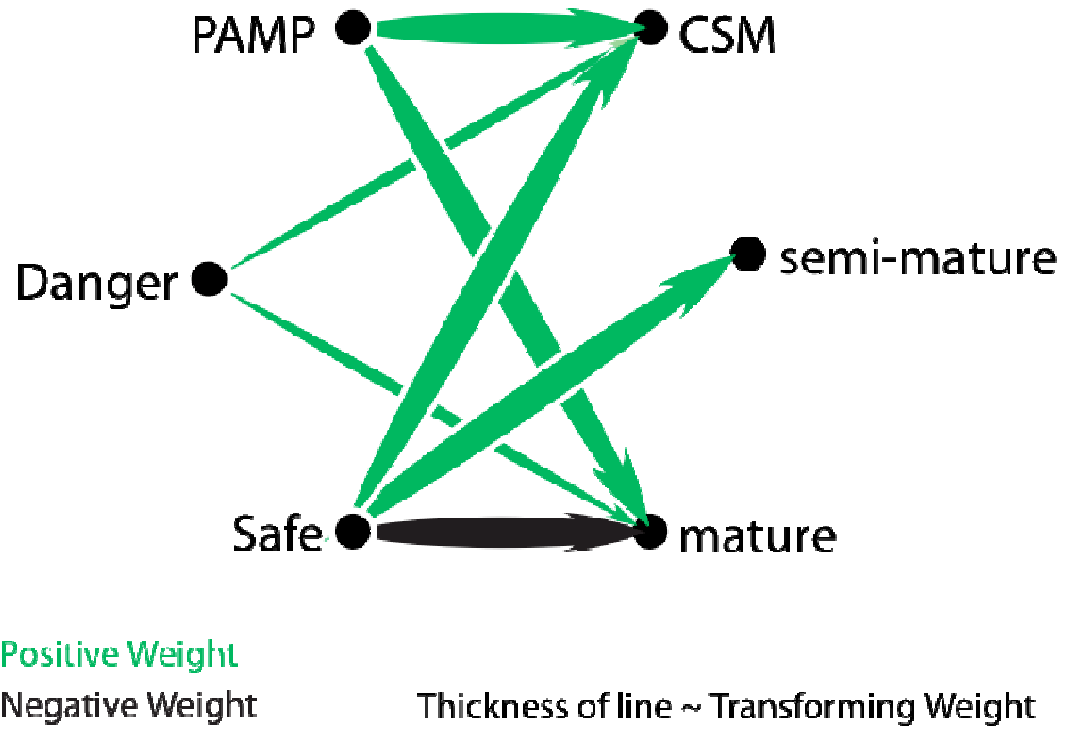

Figure 5: Input / Output signal transformation within a DC in the DCA.

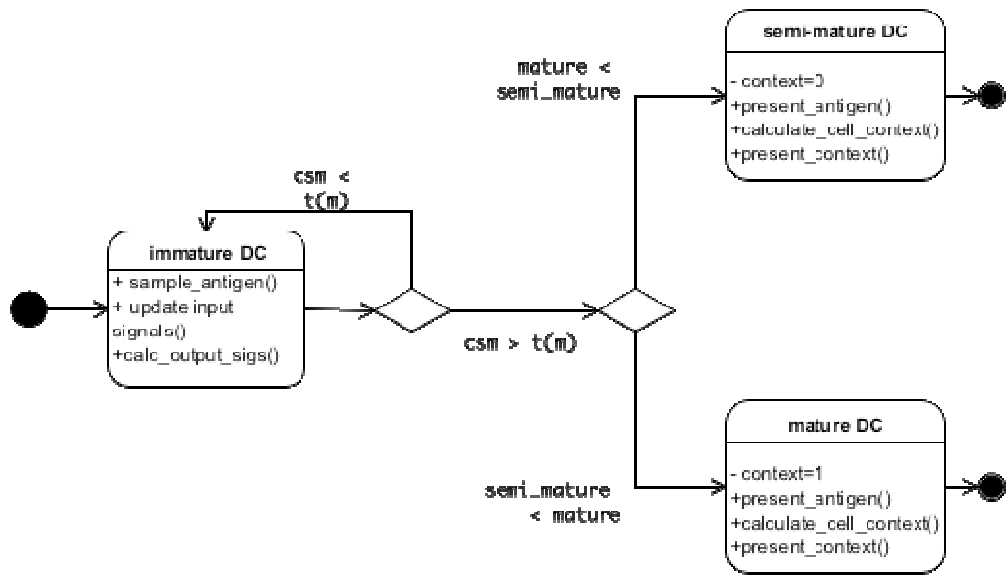

Figure 6: How immature DCs turn into either semi-mature or mature DCs in the DCA.

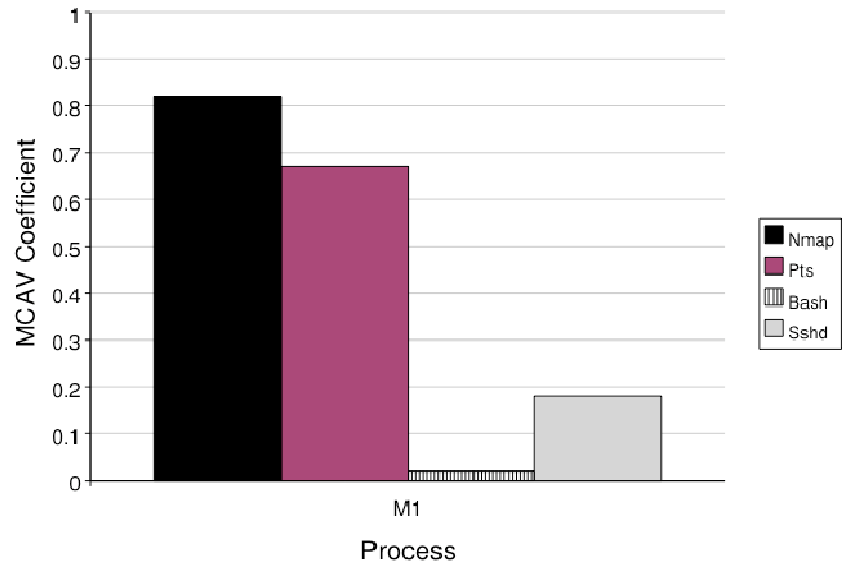

Figure 7: Results of the DCA applied to Ping Scan Detection